%% file: main.tex
\pdfoutput=1

\documentclass[11pt]{article}

\usepackage[final]{acl}

\usepackage{times}
\usepackage{latexsym}

\usepackage[T1]{fontenc}

\usepackage[utf8]{inputenc}

\usepackage{microtype}

\usepackage{inconsolata}

\usepackage{graphicx}

\usepackage{url}
\newcommand{\updates}[1]{\textcolor{black}{#1}} 
\newcommand{\cameraready}[1]{\textcolor{black}{#1}} 

\usepackage{CJKutf8}
\usepackage[utf8]{inputenc}
\usepackage[T1]{fontenc}

\newcommand{\ja}[1]{\begin{CJK}{UTF8}{ipxm}#1\end{CJK}}
\newcommand{\dataset}{\textsc{AdParaphrase}}

\usepackage{booktabs}
\usepackage{multirow}
\usepackage{amssymb}

\title{\dataset: Paraphrase Dataset for Analyzing Linguistic Features toward Generating Attractive Ad Texts}

\author{
    Soichiro Murakami$^{1}$, \ Peinan Zhang$^1$, \ Hidetaka Kamigaito$^{2,3}$, \\
    {\bf Hiroya Takamura}$^3$, \  {\bf Manabu Okumura}$^3$ \\
  $^1$CyberAgent, Inc., $^2$Nara Institute of Science and Technology, $^3$Institute of Science Tokyo \\
  {\tt 	\{murakami\_soichiro,zhang\_peinan\}@cyberagent.co.jp}, \\
  {\tt kamigaito.h@is.naist.jp}, {\tt \{takamura,oku\}@pi.titech.ac.jp} \\
}

\begin{document}
\maketitle
\begin{abstract}
Effective linguistic choices that attract potential customers play crucial roles in advertising success. 
This study aims to explore the linguistic features of ad texts that influence human preferences. 
Although the creation of attractive ad texts is an active area of research, progress in understanding the specific linguistic features that affect attractiveness is hindered by several obstacles. 
First, human preferences are complex and influenced by multiple factors, including their content, such as brand names, and their linguistic styles, making analysis challenging. 
Second, publicly available ad text datasets that include human preferences are lacking, such as ad performance metrics and human feedback, which reflect people's interests. 
To address these problems, we present \dataset, a paraphrase dataset that contains human preferences for pairs of ad texts that are semantically equivalent but differ in terms of wording and style. 
This dataset allows for preference analysis that focuses on the differences in linguistic features. 
\updates{Our analysis revealed that ad texts preferred by human judges have higher fluency, longer length, more nouns, and use of bracket symbols.}
Furthermore, we demonstrate that an ad text-generation model that considers these findings significantly improves the attractiveness of a given text.
\cameraready{
The dataset is publicly available at: \url{https://github.com/CyberAgentAILab/AdParaphrase}.\footnote{The dataset is provided under the CC BY-NC-SA 4.0 license.}
}
\end{abstract}

\section{Introduction\label{sec:introduction}}
Online advertising plays a significant role in digital marketing.
Advertising aims to attract attention, spark interest, and encourage clicks and purchases for profit.
To achieve this, writing an ad text relevant to user interests is essential; however, this alone is not sufficient.
Simultaneously, the way an ad text is written, that is, the linguistic expression of the ad text to attract attention, is crucial for the success of advertisements.
Effective linguistic choices in ad texts can influence the ease of understanding and appeal to people, potentially leading to successful outcomes, such as improved click-through rates (CTR).

\input{image/motivation_example}

In this study, we aim to explore the linguistic features of an ad text that influences human preferences with the goal of maximizing the potential success of advertisements.
Many studies have been conducted on the methods for generating attractive ad texts \cite{Hughes2019-sh,Kamigaito2021-iy,wei2022creater}.
However, progress in understanding the linguistic features that affect attractiveness, that is, human preferences, has been hindered by several obstacles.
First, human preferences are complex and are influenced by various factors, including linguistic expressions and semantic content \cite{pryzant-etal-2018-interpretable,hu-etal-2023-decipherpref}.
For instance, when analyzing human preferences for two ad texts with different content and writing styles, such as ad texts (a) and (c) in Figure \ref{fig:motivation_example}, it is difficult to determine the factors influencing preferences.
They can be motivated by content such as place names (e.g., \textit{Atami}) or writing styles such as uppercase text (e.g., \textit{BOOK NOW!}).
Second, publicly available ad text datasets containing human preference data are lacking \cite{murakami2023atgsurvey}. 
Previous studies that analyzed attractive ad texts relied on log data, such as clicks and views, that reflect human preferences \cite{pryzant-etal-2018-interpretable,murakami-etal-2022-aspect}.
However, the datasets they used were proprietary to companies for confidentiality reasons and were not publicly available. 
This restricts potential researchers to those with access to such data, hindering research on what makes ads attractive.

To explore the linguistic features that affect attractiveness by overcoming the aforementioned obstacles, we present \dataset, a novel paraphrase dataset that contains human preferences for pairs of ad texts that are semantically equivalent but differ in wording and style.
We carefully constructed the dataset by collecting semantically similar ad texts, performing paraphrase identification with five judges per pair, and collecting human preference judgments with ten judges per pair, as shown in Figure \ref{fig:motivation_example}.
The dataset allows us to focus on differences in linguistic features between individual ad texts while minimizing the impact of differences in semantic content. Thus, we can analyze human preferences centered on these features.

\updates{Through a statistical analysis of the human preferences collected in our dataset, we found that an ad text that is more fluent, longer in length, contains more nouns, and uses bracket symbols tends to be preferred by most human judges. }
Based on the findings of this analysis, we explored various methods for generating more attractive ad texts in terms of their linguistic styles. 
In the experiments, we found that more attractive ad texts could be generated by considering our findings and preference judgments as few-shot examples; however, room for improvement still exists.
We believe that our dataset will advance the understanding of linguistic features that influence human preferences in advertisements.

\section{Construction of \dataset\label{sec:paraphrase_dataset_construction}}
\updates{
We constructed \dataset, a paraphrase dataset of ad texts, to examine the influence of linguistic variations on human preferences. 
This dataset allowed us to collect and analyze human preferences by focusing on the linguistic differences between ad texts. 
Our two-step dataset construction process involves the collection of paraphrase candidates (\S\ref{sec:collecting_paraphrase_candidates}) and manual annotation for paraphrase identification (\S\ref{sec:manual_paraphrase_annotation}).}

\subsection{Collecting Paraphrase Candidates \label{sec:collecting_paraphrase_candidates}}
\dataset \ was constructed based on two publicly available Japanese ad text datasets, Ad Similarity \cite{peinan2024adtec} and CAMERA \cite{mita2024camera}.\footnote{Both are governed by the CC BY-NC-SA 4.0 license and we adhere to the intended use of both datasets.}
\updates{
We employed different strategies, as outlined below, to collect paraphrase candidates from the datasets based on their distinct formats.
Table \ref{tab:paraphrase_dataset} summarizes the breakdown of the paraphrase candidates.
We obtained a total of 1,238 candidates.
}
\input{table/paraphrase_dataset_statistics}

\paragraph{Ad Similarity}
Ad Similarity is a dataset that includes 6,332 pairs of ad texts rated on a five-point scale by three human evaluators. 
Higher scores indicate greater similarity between pairs, and vice versa.
\updates{
To collect the candidates, we set a threshold of four for the similarity score and extracted 1,088 pairs.
However, we found that several extracted pairs contained different named entities such as date and price. 
To address this issue, we applied regular expression rules, removed 382 pairs, and finally obtained 706 candidate pairs.
}

\paragraph{CAMERA}
\updates{
CAMERA is a benchmark dataset for ad text generation (ATG) tasks, in which an ad text is generated from user queries and source documents. Unlike with Ad Similarity, we created paraphrases from scratch using ad texts in the dataset as the source text. We asked two professional ad writers from an advertising agency to create a paraphrased text from a given source text. We provided two instructions: first, to rephrase its wording and style to enhance attractiveness without adding or deleting any information; and second, to limit the ad text to 15 full-width characters, which is the length limit for advertising delivery platforms such as Google Ads.
Owing to time constraints, we requested approximately 100 paraphrases from human experts, resulting in 133 paraphrases. In addition, we used large language models (LLMs) that have shown remarkable paraphrasing capabilities \cite{cegin-etal-2023-chatgpt} to generate paraphrase candidates. To enable a direct comparison between LLMs and human experts in paraphrasing, we used the same source ad texts. We employed GPT-4, GPT-3.5, and Llama2 \cite{openai2024gpt4technicalreport,touvron2023llama2openfoundation} in a zero-shot setting and provided them with the instructions identical to those of human experts. This process yielded 399 pairs. A total of 532 pairs were collected from human experts and LLMs.
We present the details of the paraphrase creation, including the instructions and model hyperparameters, in Appendices \ref{appendix:paraphrase_creation_instruction} and \ref{appendix:implementation}.
}
\subsection{Manual Annotation \label{sec:manual_paraphrase_annotation}}
\updates{
For the paraphrase candidates, we conducted a manual annotation for paraphrase identification, which is a binary classification task, to determine whether each pair of ad texts is semantically equivalent (i.e., \textit{paraphrase} or \textit{non-paraphrase}). 
We defined the paraphrasing criteria at the sentence level rather than at the word or phrase level to assess whether the two sentences conveyed the same meaning.
Annotation was performed by five workers with extensive experience in in-house advertising production.
The gold label was determined by majority votes.
For details of paraphrase identification, including quality control measures and the instructions presented to workers, please refer to Appendix \ref{appendix:paraphrase_identification}.}

\subsection{Data Analysis}
\input{image/distribution_of_jaccard_similarity}
\input{table/example_of_paraphrase_nonparaphrase}
The annotation results of the paraphrase identification are summarized in Table \ref{tab:paraphrase_dataset}. 
\updates{
Of the 1,238 candidates, 725 were paraphrases, and 513 were non-paraphrases. Ad Similarity yielded 335 paraphrases, whereas human experts and LLMs produced 390 paraphrases.
}
The average length of the ad texts is 7.1 words.
The Fleiss' kappa metric among the five workers was 0.462, indicating moderate agreement \cite{fleiss1971mns,richard1977iaa}.
\updates{
Table \ref{tab:paraphrase_example} lists examples of paraphrases and non-paraphrases. 
Several cases determined to be non-paraphrases were caused by differences in named entities, such as product names.
}

\updates{
To better understand \dataset, we calculated the Jaccard similarity to measure the lexical overlap between ad text pairs.
Figure \ref{fig:dist_of_jaccard_similarity} illustrates the distributions of the metrics for the paraphrases and non-paraphrases.
}
These results confirm that \dataset \ includes many paraphrases that are lexically different but semantically equivalent.

\section{Collection of Human Preferences\label{sec:human_preference_collection}}
\updates{
We collected human preferences through human evaluations of the attractiveness of ad texts rather than relying on log data, such as CTR, for two reasons. First, human evaluations of attractiveness are widely used to measure the quality of ad texts in the ATG field \cite{murakami2023atgsurvey}. Second, previous studies have shown that human attractiveness ratings and their predicted CTR (pCTR) are reasonably consistent, supporting the validity of using human evaluations as a proxy \cite{mita2024camera}. Therefore, we expect this approach to serve as an alternative to not releasing such data publicly.}

\subsection{Evaluation Method\label{sec:attractiveness_evaluation_method}}
We used the 725 paraphrase pairs in Table \ref{tab:paraphrase_dataset} as the evaluation set and performed pairwise comparisons to determine the ad text that was more attractive in each pair. 
Each pair was evaluated by ten human judges, who are native Japanese speakers, recruited from a crowdsourcing platform.
\updates{A skip option was provided for judges to use when they found the ad texts equally attractive.}
Appendix \ref{appendix:attractiveness_eval} provides details of the attractiveness evaluation, including the workers and interface of the evaluation tool.

\subsection{Quality Control}
\updates{
Several strategies were introduced to ensure annotation quality.
(1) To mitigate position bias in pairwise comparisons, we randomized the order of ad text pairs before presenting them to the judges.
(2) Given the subjective nature of attractiveness, we provided clear evaluation criteria based on the guidelines of \citet{Wang2021-uq}, offering examples of evaluation perspectives such as \textit{catchy}, \textit{memorable}, and \textit{easy-to-read}. 
(3) To ensure the annotation quality across crowdsourced workers of varying skill levels, we incorporated dummy questions with clear answers.
These dummy questions consisted of pairs of identical ad texts, expecting the judges to select a skip option.
We rejected all responses from judges who failed to choose the skip option for the dummy questions.
}

\subsection{Evaluation Results\label{sec:attractiveness_evaluation_results}}
\input{image/max_number_of_votes}
\updates{
Figure \ref{fig:dist_of_max_number_votes} shows a histogram of attractiveness evaluations from ten judges for ad text pairs. The x-axis indicates the maximum number of judges who selected the same ad text in a pair. For instance, a value of six indicates that six judges preferred the same ad text, whereas zero indicates that all judges skipped it, suggesting equal attractiveness of the ad text pair. The distribution shows many cases in which five to six judges selected the same ad text, suggesting inconsistent preferences for paraphrased ad texts. The inter-rater agreement measured by Fleiss' kappa was 0.161, indicating a slight agreement.
}

Nevertheless, it should be noted that in 316 cases, seven or more judges agreed on their preference. 
In these cases, the inter-rater agreement was 0.307, indicating fair agreement.\footnote{\cameraready{It is worth noting that the relatively low inter-annotator agreement for ``attractiveness'' can be attributed to the subjective nature of this evaluation criterion, as also reported in previous work \cite{mita2024camera}.}}
These findings suggest that linguistic differences between paraphrased ad texts influence the preferences of human judges to a certain extent.

\section{Experiments}
According to \S\ref{sec:attractiveness_evaluation_results}, the human preference judgments showed fair agreement in 316 evaluation cases.
This raises the following question: \textit{What linguistic differences between ad text pairs affect human preference judgment?}
\updates{
To answer this question, we analyzed the linguistic features influencing human preferences, focusing particularly on cases in which the judges' preferences were aligned (\S\ref{sec:feature_analysis}). 
Furthermore, to demonstrate the practical applications of the findings, we explored methods for generating attractive ad texts that incorporated the linguistic features identified in the analysis (\S\ref{sec:attractive_ad_text_generation}).
}

\subsection{Analyzing Linguistic Features that Influence Human Preferences \label{sec:feature_analysis}}

\subsubsection{Linguistic Features}

\input{table/resulf_of_linguistic_feature_analysis}
\updates{
An ideal ad text is designed to capture attention and generate interest.
Therefore, readability, visibility, and informativeness are key factors in improving the attractiveness of ad texts \cite{hsuehcheng_wang__2012,murakami2023atgsurvey}. 
We analyzed various linguistic features from the surface-level, such as text length, to deeper factors, such as lexical choice and emotion. 
Our feature set consists of four groups: raw text, lexical, syntactic, and stylistic features, including the 24 types listed in Table \ref{tab:chi_square_test}.
For the detailed definitions of these features, see Appendix \ref{appendix:linguistic_features}.}
\updates{
\paragraph{Raw text features}
Raw text features, such as text length, are the most basic features of ad text that affect readability and informativeness. 
We measured word and character counts.}
\updates{
\paragraph{Lexical features}
Lexical features include word-level features such as the number of content words, lexical choice, and character types.
We hypothesized that ad texts with more content words would be more informative and attractive.
Regarding lexical choices, we posit that ad texts using common words would be more readable and preferable. 
To measure this, we calculated the average word frequency of each ad text using the Balanced Corpus of Contemporary Written Japanese \cite{maekawa-etal-2010-design} and counted the common and proper nouns. 
Because \dataset \ comprises Japanese ad texts, we also included character-type features, such as hiragana counts, which are known to affect readability by helping readers distinguish words more clearly \cite{sato-etal-2008-automatic}.}
\updates{
\paragraph{Syntactic features}
Syntactic features relate to the structure of the entire ad text or its parts, such as phrases, including the depth of the dependency tree, length of the dependency links, number of noun phrases, and perplexity (PPL).}
\updates{
\paragraph{Stylistic features}
Stylistic features comprise features related to the wording and style of ad texts, including high-level linguistic features such as emotions (e.g., joy and anticipation) and the specificity of the ad text.
We hypothesized that more positive or specific ad texts would be preferred.
To determine these features, we used separate classification models for emotion and specificity, which we had constructed (see Appendix \ref{appendix:linguistic_features} for details).
In addition, we used the presence or absence of bracket symbols such as \ja{【】} or \ja{「」} as a unique linguistic feature in Japanese ad texts because these brackets are often used to highlight important information for visibility. 
For instance, in the ad text \textit{\ja{【Official Site】ABC Insurance}}, the brackets highlight the key information.
}

\subsubsection{Analysis Method \label{sec:linguistic_feature_analysis_method}}
\updates{
We used the chi-square test of independence to analyze the relationship between the linguistic features and human preferences. This statistical method tests the independence of two categorical variables. In our study, these variables were (1) the ad text selected by the majority of human judges and (2) the ad text scoring value (higher or lower) for each linguistic feature. For example, when examining the impact of PPL on human preferences, we investigated whether ad texts with lower PPL tended to be selected by the majority of the judges.
}

\updates{
To ensure the reliability of our analysis, we used the 316 (out of 725) pairs of ad texts, in which the choices of more than seven human judges for attractiveness were consistent, as this indicated fair agreement by Fleiss' kappa \cite{richard1977iaa}.
}

\subsubsection{Results\label{sec:feature_analysis_results}}
\updates{
Table \ref{tab:chi_square_test} lists the chi-square test results. Our analysis focused on the differences in features between the two ad texts, considering only the cases in which each feature differed within the pair. Consequently, the number of cases varied for each feature type. For example, the character counts differed in 248 pairs.
}

\updates{
In Table \ref{tab:chi_square_test}, linguistic features with high chi-square values and low p-values indicate a strong correlation with human preferences.
Therefore, the results demonstrate that the ad texts preferred by the majority of human judges exhibit significantly lower PPL, longer character counts, higher frequencies of nouns, more noun phrases, and greater use of bracket symbols ($p<0.01$).
These findings suggest that to create attractive ad texts, it is crucial to focus on these specific linguistic features.
}

\updates{
However, we observed no statistically significant differences in other linguistic features.
Several factors may explain these results.
We speculated that the complexity of the dependency trees had little impact because of the short length of the ad texts.
The quality of emotion labels may be limited by a domain mismatch because the emotion classification model was trained on social media texts, which differ from ad texts.
Concerning specificity, because we used paraphrase pairs for attractiveness evaluation, the specificity of the pair differed in a few cases and we could not obtain a sufficient number of cases for analysis. 
However, a promising result was obtained, in that 10 out of the 11 cases with high specificity were preferred by the majority of human judges.
}

\subsection{Generating Attractive Ad Texts \label{sec:attractive_ad_text_generation}}
\updates{
To demonstrate the practical application of these findings, we conducted an experiment using ATG. Specifically, we explored methods for generating attractive ad texts based on the findings of our analysis.\footnote{\updates{In this experiment, we tested the findings other than the frequency of nouns and noun phrases.}}
}

\subsubsection{Ad Text Generation Methods\label{sec:atg_generation_methods}}
In this experiment, we focused on ad text refinement \cite{youngmann2020,mishra2020refinement}.
The goal was to generate more attractive ad text by rephrasing its wording and style without adding or deleting information from the input ad text.

\updates{
Inspired by the success of integrating human preferences into LLMs, we explored methods for generating attractive ad texts using LLMs. 
Various methods have emerged for integrating human preferences into LLMs, including reinforcement learning, supervised fine-tuning, and in-context learning (ICL) \cite{ouyang_instructGPT_2022,patrick_human_feedback_survey_tacl,kirk-etal-2023-past}.
Owing to limited preference data, we focused on ICL, which is a learning paradigm in which an LLM learns a new task from context, including instructions and input--output demonstrations, and is known for its effectiveness in few-shot settings \cite{brown_gpt3_2020}.}

\updates{
We explored methods for generating more attractive ad texts by learning human preferences using ICL. For instructions, we incorporated the findings of the linguistic features into the prompt. For the demonstrations, we created two types of input--output pairs using the results of attractiveness evaluations: those in which attractiveness improved and those in which it did not. 
For the former, we selected cases in which the preferences of at least seven out of ten human judges matched, using the less-preferred text as the input and the more-preferred text as the output. For the latter, we selected cases in which no consensus was reached among the judges. These were referred to as positive and negative examples, respectively. Full experimental details, including prompts, are provided in Appendices \ref{appendix:ad_text_generation} and \ref{appendix:implementation}.
}

\input{table/result_of_ad_text_generation}

\subsubsection{Evaluation Method\label{sec:atg_evaluation_method}}
We conducted a human evaluation based on two aspects: paraphrase identification\footnote{\updates{Although automatic metrics such as ROUGE can be used, \citet{shen-etal-2022-evaluation} reported that they do not align well with human judgments. Therefore, we opted for a human evaluation.}} and attractiveness. Specifically, we assessed whether the generated ad texts were semantically equivalent to the input ad texts and which were more attractive, following the same procedures in \S\ref{sec:manual_paraphrase_annotation} and \S\ref{sec:attractiveness_evaluation_method}.

To verify the effectiveness of the instructions and demonstrations, we compared five variations of GPT-4-based models listed in Table \ref{tab:atg_result}, in addition to the zero-shot baselines of Llama2, GPT-3.5, and GPT-4 in \S\ref{sec:paraphrase_dataset_construction}.
For example, \textrm{GPT-4-fewshot-findings-pos} refers to the GPT-4 model that uses the findings and positive examples as few-shot examples.
For comparison with the baselines, we used the same 133 ad texts as inputs for the evaluation.
To prevent data leakage, the ad texts were not used as few-shot examples.

\subsubsection{Results and Discussion\label{sec:atg_evaluation_results}}
Table \ref{tab:atg_result} summarizes the results. The success rate for each evaluation task was calculated. For paraphrase identification, the success rate represents the proportion of generated texts labeled as paraphrases by the majority of human judges. For attractiveness, we defined the success rate as the proportion of paraphrased texts found to be more attractive by at least seven out of ten judges. We also calculated the overall success rates for both tasks.
The following can be observed from Table \ref{tab:atg_result}.

\paragraph{Findings significantly improved performance}
Explicitly, incorporating the findings into the instructions (e.g., \textrm{GPT-4-zeroshot-findings}) significantly improved both evaluation tasks compared to the baselines (e.g., \textrm{GPT-4-zeroshot}). We found that directly incorporating these findings into the instructions was straightforward and effective. Further improvements would be possible by identifying other factors that influence preferences.

\paragraph{Negative examples are also useful}
\input{table/jaccard_similarity_between_input_and_generated_ad_texts}
Introducing only positive examples as few-shot examples improves the paraphrase identification performance. However, contrary to our expectations, this yielded little improvement in attractiveness. 
Interestingly, when we introduced negative examples as well, the attractiveness evaluation performance improved significantly.\footnote{\cameraready{Fisher's exact test showed GPT-4-fewshot-both significantly outperformed GPT-4-fewshot-pos (p < 0.05), while no significant difference was found between GPT-4-fewshot-findings-pos and -both.}}
One possible explanation is that negative examples consist of minor edits between the input and output texts compared to positive examples. For example, the average Jaccard similarity between the input and output were 0.560 and 0.602 for the positive and negative examples, respectively. 
This discrepancy may have influenced the generation model to actively perform editing operations such as the active use of brackets.
\cameraready{
To verify this, we measured the Jaccard similarity between the input and generated text for models using only positive examples and those using both positive and negative examples. 
The results are shown in Table \ref{tab:jaccard_similarity_between_input_and_generated_ad_text}.
Lower values indicate less lexical similarity between the input and generated text, suggesting that the output has been actively rephrased. These results further support the explanation.
}
These findings suggest that considering both positive and negative examples can help the model learn the characteristics of attractive ad texts.

\paragraph{Attractiveness has room for improvement}
According to the overall success rate, the model (\textrm{GPT-4-fewshot-findings-both}), which incorporated the findings and both positive and negative examples, performed the best, outperforming HUMAN.
However, room for improvement still exists in terms of attractiveness, even though the five models performed well in paraphrase identification.
Therefore, it is worth exploring the factors that influence human preferences and the methods for generating more attractive ad texts in future work.

\input{table/analysis_of_generated_ad_texts}
To examine whether the generated texts reflected the linguistic features provided by ICL, we analyzed the linguistic features of the generated texts. 
Specifically, we calculated the features identified in the analysis, including PPL, number of characters, and presence of brackets. 
Table \ref{tab:atg_analysis} presents the results.
These results confirm that \textrm{GPT-4-fewshot-findings-both}, the model with the highest overall success rate, tends to have a lower PPL, longer text length in characters, and a higher proportion of texts containing brackets, indicating that the generated texts successfully reflect the findings.
\cameraready{
Importantly, the results in Table \ref{tab:atg_result} and Table \ref{tab:atg_analysis} suggest that the linguistic features identified through the linguistic feature analysis (\S\ref{sec:feature_analysis}) contribute to enhancing the attractiveness of ad texts.
}

\section{Analysis\label{sec:analysis}}

\updates{
Through experiments, we explored the linguistic features that influence human preferences and the methods for generating attractive ad texts.
Because the primary goal of advertising is to capture users' attention, it is crucial to examine how attractively paraphrasing ad text affects ad performance, such as clicks. 
To this end, we analyzed the relationship between human preference and pCTR. 
In addition, we conducted an online evaluation of paraphrased ads in real-world ad-delivery scenarios.}

\subsection{Relationship between Human Preferences and pCTR}
\input{table/alignment_of_preference_and_pctr}
\updates{
Table \ref{tab:alignment_preferece_pctr} lists the number of cases in which human preferences align with pCTR in the evaluation set. 
Following \citet{Hughes2019-sh}, we used an in-house CTR prediction model.
We identified ad texts that were both preferred and had a higher pCTR. Only 355 of the 725 ad texts exhibited this alignment, suggesting a weak correlation between human preference and pCTR.
For a detailed analysis, we divided the evaluation data into two groups based on the level of agreement. The high-agreement group, in which more than seven judges preferred the same ad, showed a 54.7\% alignment, while the low-agreement group, which included all other cases, showed a 44.5\% alignment. This suggests that the ad texts preferred by many judges tend to align with pCTR. Therefore, attractive rephrasing can positively affect pCTR, although to a limited extent.}

\subsection{Ad Performance via Online Evaluation}
\input{table/result_of_online_evaluation}
We conducted an online evaluation to verify the extent to which the paraphrasing method (\S\ref{sec:attractive_ad_text_generation}) affected the ad performance. 
Specifically, we performed an A/B test to compare the performance of the original ads as a baseline with that of the paraphrased ads. 
\cameraready{
See the link\footnote{\url{https://support.google.com/google-ads/topic/3121777}} for the glossary of advertising terms.
We used Google Ads' search ads as the evaluation platform.
For the evaluation data, we used three ad groups from two companies in the human resources (HR) and education industries. 
Each ad group consists of a maximum of 15 headlines and four descriptions.
We applied the paraphrasing methods to the headline ad text.
}
We delivered the original and generated ads for two weeks and compared the number of impressions, clicks, and costs, which is the budget spent. 
The higher the value, the better for all metrics.

Table \ref{table:online_evaluation} presents the results of the online evaluation, indicating a relative improvement in the generated ads compared to the original ads. 
A value exceeding 1 indicates the superior performance of the generated ad.\footnote{Due to confidentiality, absolute values for each metric cannot be disclosed.} 
For example, if the baseline impressions were 100 and the generated text had 110, the improvement ratio will be $110/100=1.1$.
According to Table \ref{table:online_evaluation}, performance varies across ad groups; the ad performance for the HR company improved, whereas the others degraded.
Future work should focus on conducting extensive online evaluations and developing paraphrasing methods that consistently enhance performance across diverse ad texts.

\section{Discussion for Practical Application}
\cameraready{
As a potential practical application of this study, the development of a writing assistant tool based on our paraphrasing method could be considered to enhance the attractiveness of input ad texts. 
In advertising production, there are situations where the advertised content is fixed due to campaign requirements, and only the expression of the ad text needs to be refined.
In fact, many tools already exist with the aim of supporting ad production in various ways.\footnote{\url{https://cyberagent.ai/products/}}
Such a writing assistant tool could further improve the wording and style of ad texts, making them more appealing.
}

\section{Related Work}
\subsection{Generating Attractive Ad Text\label{sec:related_work_atg}}
With the increasing demand for online advertising, the manual creation of ad text has reached its capacity limits. 
Therefore, researchers have focused on ATG \cite{murakami2023atgsurvey}.
The goal of advertising is to attract interest in a product or service and motivate users to take action such as clicking and purchasing. 
Therefore, generating attractive ads is critical to the success of online advertising.

\updates{
Various methods have been developed to generate attractive ad texts, ranging from template-based approaches \cite{Bartz2008-ke,thomaidou2013} to neural-based techniques \cite{Hughes2019-sh,Kamigaito2021-iy}.
Our work differs from previous studies in that it focuses on how the wording and style of ad text affect attractiveness. 
Unlike previous studies that evaluated generated ad texts without considering content variations, we ensured semantic equivalence between ad pairs before assessing their attractiveness. 
This approach allowed us to specifically analyze the impact of expressions on attractiveness. 
We believe that identifying the key linguistic factors that influence the attractiveness of ad texts is crucial for exploring the potential directions for advancing ATG methods.
}

\subsection{Understanding Attractive Ad Text}
Understanding the factors that affect the attractiveness of ad text is crucial to the success of the ad creation process. 
Various efforts have been made to analyze the factors influencing the attractiveness of ad texts, such as advertising appeal \cite{murakami-etal-2022-aspect}, persuasive tactics \cite{yuan2023persuadetoclick}, and emotions \cite{youngmann2020}. 
This study investigates the linguistic features of ad texts that affect their attractiveness.

\updates{
A common approach for studying the factors influencing human preferences is to use log data, which measures attractiveness based on clicks and views. However, these log data are often proprietary and not publicly available, hindering research replication and knowledge advancement.
Therefore, we collected human preferences through manual evaluations to make the data publicly available.
}

In a study closely related to our work, \citet{pryzant-etal-2018-interpretable} examined the impact of writing style on ad performance, while controlling for potential confounding variables. 
However, this study only considered the influence of the brand names and neglected other content-related factors. 
In our study, we constructed a paraphrase dataset varying in writing style to focus on the linguistic differences between ad texts while mitigating content variations, thereby enabling the analysis of human preferences centered on these linguistic features.

\section{Conclusion}
In this paper, we proposed a dataset called \dataset, which comprises semantically equivalent ad pairs, to analyze the linguistic features that influence human preferences.
We collected data on human preferences for these ad pairs and demonstrated that factors such as fluency, number of characters, and use of brackets significantly affected the attractiveness of the ad text. 
\updates{
In addition, we conducted experiments on ATG, showing that considering human preferences can lead to the generation of attractive ad texts.
In future work, we plan to expand our dataset to analyze other attractiveness factors and explore ATG methods.
}

\section{Limitations\label{sec:limitations}}
This study had several limitations that should be addressed in future studies.
\updates{
\paragraph{Dataset size} \dataset \ consists of 1,238 pairs of ad text, including 725 paraphrases. 
Although the limited size of the dataset has affected the robustness and generalizability of our findings, we ensured reliable results. 
Unlike many existing studies that use three human judges per pair to evaluate attractiveness, we asked ten human judges per pair to collect more comprehensive preference data. 
This approach helped mitigate the limitations of dataset size. 
In future work, we plan to expand the dataset to include a broader range of linguistic features in ad texts, thus further enhancing the depth and applicability of our research.}
\updates{
\paragraph{Linguistic features}
In this study, we analyzed a variety of linguistic features, including deep linguistic features that extend beyond surface-level features such as lexical choice and textual specificity. However, other unexplored features such as discourse intent were not considered. Therefore, future studies should consider these additional linguistic features to provide a more comprehensive understanding of the factors that influence human preferences.
}
\updates{
\paragraph{Generalizability to other languages}
\dataset \ consists of Japanese ad texts, and all linguistic features and preference analyses are based on Japanese ad texts. 
Therefore, please note that some features, such as character types, are unique to the Japanese language, and we do not intend that our analysis results be applicable to all languages.
However, we believe that other languages such as English and Chinese also have their unique linguistic features.
Multilingualization of a dataset is an important future direction for revealing linguistic features that can be applied to a wide range of languages.
In this study, we focused on Japanese ad texts, but we hope that this will pave the way for the development of advertisement paraphrasing data in different languages.
}

\paragraph{Annotator demographics}
Although human preferences can be influenced by the demographic background of the annotators, we did not collect such information due to constraints within the crowdsourcing platform.
Consequently, we did not consider the effects of demographic factors.
Therefore, future studies should include an analysis of preferences that considers the demographic attributes of annotators to understand how these factors influence preference judgments.

\bibliography{main}

\appendix

\section{Details of Paraphrase Creation\label{appendix:paraphrase_creation_instruction}}
We sampled the ad texts from the dev set of CAMERA \cite{mita2024camera}, a publicly available dataset for ATG tasks. 

The process of creating paraphrased candidates using CAMERA was divided into two steps.
In the first step, we asked two human experts working as professional ad writers in an advertising agency to rephrase their wording or styles to enhance the attractiveness of the given ad text.
We instructed ad writers under the following two conditions:
The first was not adding new information or deleting any existing information from the text.
Second, the length of the ad text was limited to 15 full-width characters, which is the length limit for advertising delivery platforms such as Google Ads\footnote{\url{https://support.google.com/google-ads/answer/1704389}}.
Figure \ref{fig:instruction_for_paraphrase_creation_human} shows the instructions presented to the human experts for creating paraphrases.
Due to their limited working hours, we asked them to create approximately 100 paraphrases, which resulted in 133 paraphrases.
\input{image/instruction_for_paraprhase_creation_human}

In the second step, we used LLMs including GPT-3.5, GPT-4, and Llama2 to generate paraphrased ad texts from the source ad texts. 
Figure \ref{fig:prompt_for_paraphrase_creation_llm} shows the prompt for the paraphrase creation.
A total of 133 source ad texts, identical to those presented to human experts, were used as inputs.
This was performed to allow further analysis to directly compare the abilities of LLMs and human experts to generate paraphrases.

\input{image/prompt_for_paraprhase_creation_llms}

\updates{\section{Details of Paraphrase Identification\label{appendix:paraphrase_identification}}}
\input{image/paraphrase_identification}
\input{image/paraphrase_and_non_paraphrase_examples_for_paraphrase_identification}
For the paraphrase candidates collected in \S\ref{sec:collecting_paraphrase_candidates}, we conducted a manual annotation for paraphrase identification as a binary classification task to determine whether a pair was a \textit{paraphrase} or \textit{non-paraphrase}. Our defined criteria for paraphrases involved determining whether each candidate pair of ad texts was semantically equivalent at the sentence level rather than at the word or phrase level. The annotations were performed by five in-house workers with extensive experience in advertising creation. To ensure a consistent understanding of the annotation criteria and maintain quality control, we conducted two practice sessions in advance and provided comprehensive feedback on all the questions. The final gold label was determined by a majority vote of the five workers. The user interface and the instructions presented to the judges are presented in Figure \ref{fig:user_interface_paraphrase_identification}, \cameraready{while the annotation examples presented to the judges are shown in Figure \ref{fig:annotation_example_for_paraphrase_identification}.}

\section{Details of Attractiveness Evaluation\label{appendix:attractiveness_eval}}
\input{image/attractiveness_evaluation}
The attractiveness of each ad pair was evaluated by ten workers. Among them, four were workers with experience in in-house dataset production and the remaining six were workers from a crowdsourcing platform. We used Yahoo! Crowdsourcing\footnote{\url{https://crowdsourcing.yahoo.co.jp/}} as the crowdsourcing platform. The compensation was set at 10 yen per 15 tasks, based on the platform's regulations. They were informed that the annotated results would be used for research purposes. The user interface and instructions presented to judges are shown in Figure \ref{fig:user_interface_attractiveness_eval}.

\updates{\section{Details of Linguistic Features \label{appendix:linguistic_features}}}
This study analyzed a wide range of linguistic features, from surface-level features to deeper factors, including lexical choice and textual specificity. Our feature set consists of four groups: raw text, lexical, syntactic, and stylistic features, including the 24 types of linguistic features listed in Table \ref{tab:chi_square_test}.

In the following sections, we explain the definitions and extraction methods for each feature.

\subsection{Raw text features}
\paragraph{Text length}
We calculated the number of words and characters as a measure of text length. 
We used Sudachi \cite{takaoka2018sudachi}, a Japanese morphological analyzer, to split Japanese texts into words.

\subsection{Lexical features}
\paragraph{Content words}
We hypothesized that ad texts with more content words would be more informative and attractive.
We defined content words as nouns, verbs, adjectives, adjectival nouns, and adverbs and counted their occurrences in each ad text. 
For morphological analysis of Japanese ad texts, we used Sudachi \cite{takaoka2018sudachi}. 

\paragraph{Lexical choice}
For word frequency, we hypothesized that ad texts containing more common words would be preferred; therefore, we calculated the frequency of words in each ad text. Specifically, we calculated the average frequency of occurrence per million words for each word in the ad text using the Balanced Corpus of Contemporary Written Japanese \cite{maekawa-etal-2010-design}. Furthermore, to verify whether ad texts containing common or proper nouns were preferred, we used Sudachi to extract these nouns and counted their occurrences.

\paragraph{Character type}
We counted the numbers of hiragana, katakana, kanji, symbols, and numerals in each Japanese ad text.

\subsection{Syntactic features}

\paragraph{Dependency tree}
We used spaCy with GiNZA\footnote{\url{https://github.com/megagonlabs/ginza}} to perform the dependency parsing of ad texts. The depth of the dependency tree refers to the longest path from the roots to leaves in the dependency tree. The length of the dependency link refers to the number of words between the syntactic head and its dependent.

\paragraph{Noun phrases}
We used spaCy with GiNZA to extract the noun phrases and counted their occurrences.

\paragraph{Perplexity} 
We calculated the perplexity using GPT-2 \footnote{\url{https://huggingface.co/rinna/japanese-gpt2-medium}} \cite{radford2019language} trained on a web-crawled corpus and the Wikipedia dataset.

\subsection{Stylistic features}

\paragraph{Emotion}
To label the emotion of ad texts, we used the LUKE model\footnote{\url{https://huggingface.co/Mizuiro-sakura/luke-japanese-large-sentiment-analysis-wrime}} \cite{yamada-etal-2020-luke} trained on WRIME \cite{kajiwara-etal-2021-wrime}, which is a Japanese emotion analysis dataset based on social media text. This model is an 8-class classifier that determines the most appropriate emotions from the following eight categories: joy, sadness, anticipation, surprise, anger, fear, disgust, and trust. When we applied the classifier to the ad texts, the majority of cases were labeled as either ``joy'' or ``anticipation,'' so we used only these two labels in our analysis. The accuracy of the classifier was 68.6\%.

\paragraph{Textual specificity}
Owing to the lack of datasets or existing models for measuring the specificity of Japanese sentences, we created our own specificity classifier using GPT-4 with a few-shot setting. We defined this as a three-class classification problem and constructed a model that compares two ad pairs and judges with higher specificity. If the specificity of both is equivalent, it outputs a label of ``equal.'' To evaluate the performance of the model, we randomly sampled 100 predictions and conducted a manual evaluation, achieving an accuracy of 88.0\%.

\paragraph{Brackets}
We created a binary label indicating whether the ad text contains bracket symbols \ja{【】} or \ja{「」}.

\section{Details of Ad Text Generation Experiment\label{appendix:ad_text_generation}}
\input{image/prompt_for_ad_text_generation}
Figure \ref{fig:prompt_for_ad_text_generation} shows an example of the prompts used in the ATG experiment. Specifically, this prompt was used for \textrm{GPT-4-fewshot-findings-both} in Table \ref{tab:atg_result}. In this prompt, we include few-shot examples and the findings revealed from the analysis of linguistic features that influence human preferences. For the few-shot examples, we created two types of examples, where attractiveness improved and where it did not, from the results of the attractiveness evaluation and used 20 examples for each type. We report the results of a single experiment.

\section{Implementation Details \label{appendix:implementation}}

\paragraph{Models}
We used GPT-3.5, GPT-4, and Llama2 to perform the paraphrase candidate generation (\S\ref{sec:collecting_paraphrase_candidates}) and an experiment using ATG (\S\ref{sec:attractive_ad_text_generation}). For GPT-3.5 and GPT-4, we used the Azure OpenAI API\footnote{\url{https://azure.microsoft.com/en-us/products/ai-services/openai-service/}} service with version \texttt{2024-03-01-preview} and a default temperature parameter of 1.0. In the case of Llama2, we used the \texttt{ELYZA-japanese-Llama-2-7b} model available via Hugging Face\footnote{\url{https://huggingface.co/}}, which enhances Japanese-language capabilities through additional pre-training based on Llama2. See link\footnote{\url{https://huggingface.co/elyza/ELYZA-japanese-Llama-2-7b}} for the detailed parameters of the Llama2 model.

\paragraph{Tokenizer}
We used Sudachi \cite{takaoka2018sudachi}, a Japanese morphological analyzer, to tokenize Japanese text into words. Sudachi offers three splitting modes to provide tokens of different granularities for each application purpose: short, medium, and named entities. In this study, we used the named entity mode.

\end{document}

%% file: image/motivation_example.tex
\begin{figure}[t]
 \centering
  \includegraphics[width=1\linewidth]{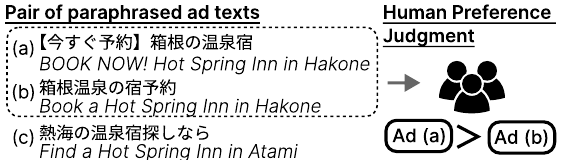}
 \caption{Overview of human preference judgments for a pair of paraphrase ad texts.}
 \label{fig:motivation_example}
\end{figure}

%% file: table/paraphrase_dataset_statistics.tex
\begin{table}[t]
\centering
{\small
\begin{tabular}{@{}l@{}l@{}r@{}r@{}r@{}} \toprule
\multicolumn{1}{c}{\textbf{Source}} & 
\multicolumn{1}{c}{\textbf{Model}} & 
\multicolumn{1}{c}{\textbf{\#Cand.}} & 
\multicolumn{1}{c}{\textbf{\#Para.}} & 
\multicolumn{1}{c}{\textbf{\#Non-para.}} \\ \midrule
Ad Similarity & \multicolumn{1}{c}{$-$} & 706 & 335 & 371 \\
\multirow{4}{*}{CAMERA} & Llama2 & 133 & 86 & 47 \\
& GPT-3.5 & 133 & 98 & 35 \\
& GPT-4 & 133 & 81 & 52 \\
& Human & 133 & 125 & 8 \\ \midrule
& \multicolumn{1}{c}{\textbf{Total}} & 1,238 & 725 & 513 \\ \bottomrule
\end{tabular}}
\caption{Statistics of \dataset. \#Cand. is the number of paraphrase candidates. \#Para and \#Non-Para denote the number of candidates identified as paraphrase and non-paraphrase, respectively.}\label{tab:paraphrase_dataset}
\end{table}

%% file: image/distribution_of_jaccard_similarity.tex
\begin{figure}[t]
 \centering
  \includegraphics[width=1\linewidth]{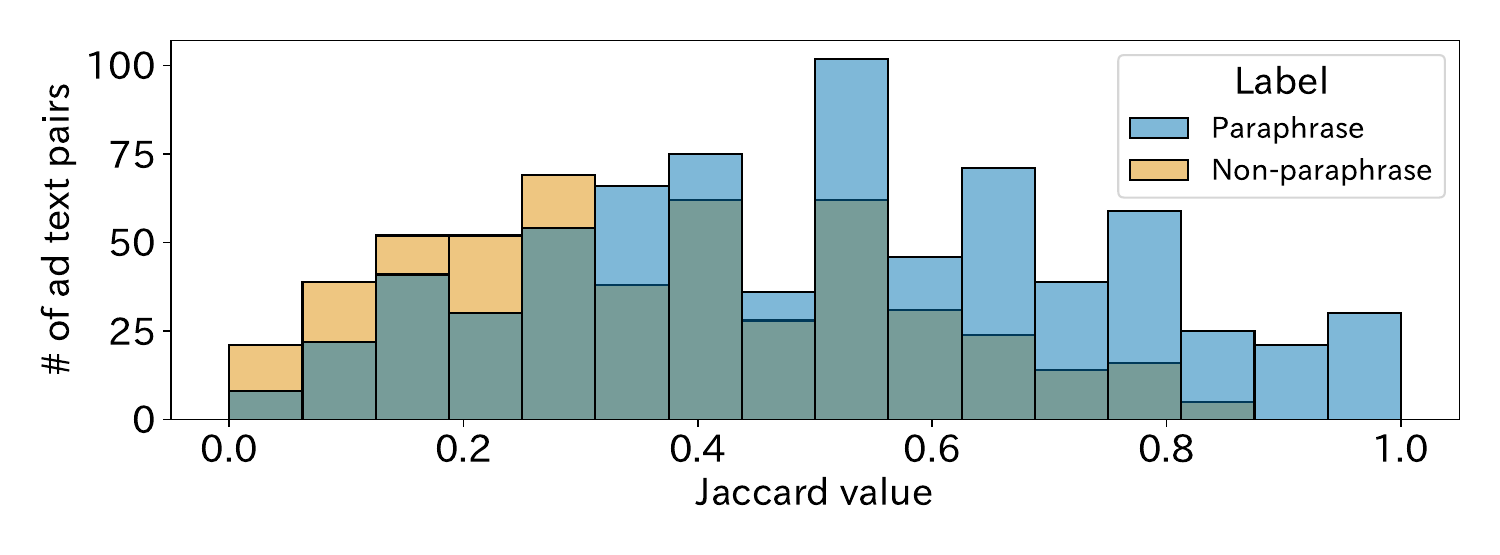}
 \caption{Distribution of Jaccard similarity for paraphrases and non-paraphrases in \dataset.}
 \label{fig:dist_of_jaccard_similarity}
\end{figure}

%% file: table/example_of_paraphrase_nonparaphrase.tex
\begin{table}[t]
\centering
{\small
\begin{tabular}{@{}l@{\ }l@{}}
\toprule
\multirow{4}{*}{\textbf{Paraphrase}} 
& \ja{「リピート確定」の声多数} \\
& \textcolor{black!70}{(\textit{Many Voices: ``Will Definitely Repeat''})} \\
& \ja{これからも使い続けたいの声続出} \\
& \textcolor{black!70}{(\textit{Many Voices Want to Keep Using It})} \\
\midrule
\multirow{4}{*}{\textbf{Non-paraprhase}}
& \ja{最強スペックのAndroid端末} \\
& \textcolor{black!70}{(\textit{The Most Powerful Android Phone})} \\
& \ja{Google Pixel 史上最強スペック} \\
& \textcolor{black!70}{(\textit{The Most Powerful Google Pixel Ever})} \\
\bottomrule
\end{tabular}}
\caption{Examples of paraphrases and non-paraphrases.}\label{tab:paraphrase_example}
\end{table}




%% file: image/max_number_of_votes.tex
\begin{figure}[t]
 \centering
  \includegraphics[width=1\linewidth]{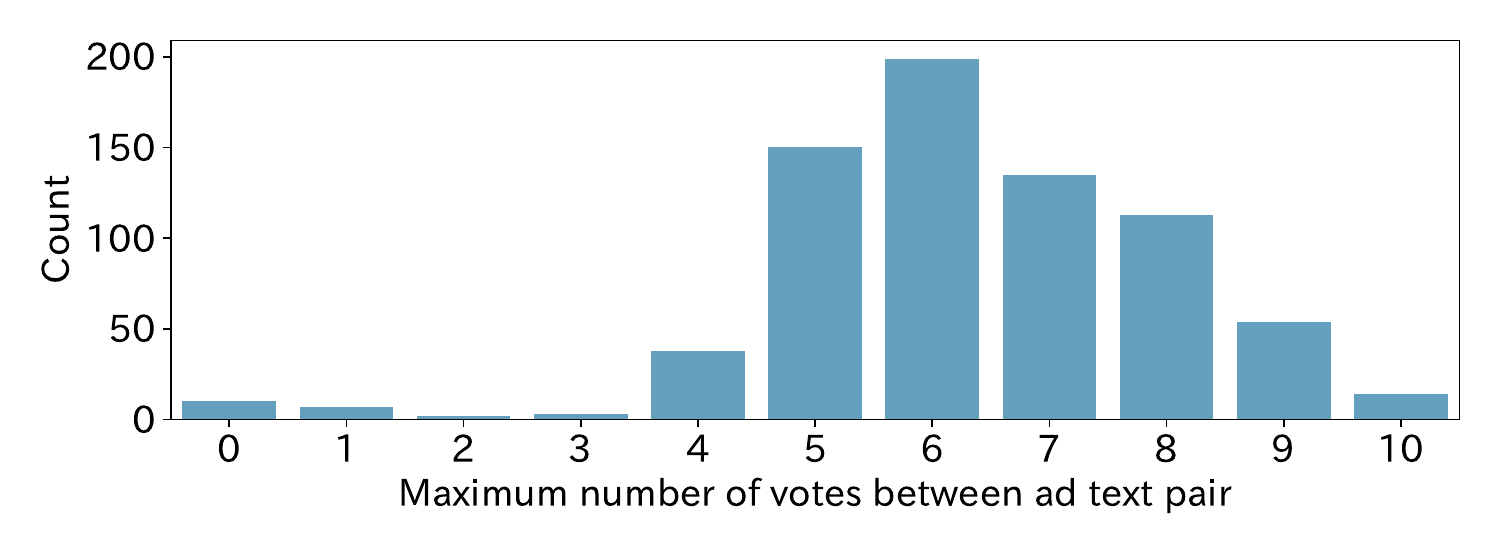}
 \caption{Distribution of maximum number of votes between ad text pair in attractiveness evaluation.}
 \label{fig:dist_of_max_number_votes}
\end{figure}

%% file: table/resulf_of_linguistic_feature_analysis.tex

\begin{table}[t]
\centering
{\small
\begin{tabular}{@{}c@{\hspace{0.2em}}l@{\hspace{0.2em}}r@{\hspace{0.5em}}r@{\hspace{0.5em}}r@{\hspace{0.5em}}r@{}}
\toprule
\multicolumn{2}{c}{\textbf{Features}} & \textbf{df} & \textbf{N} & \textbf{$\chi^2$} & \textbf{p-value} \\ \midrule
\multirow{3}{*}{\shortstack{Raw text\\features}}  
    & \textit{Text length} & & & & \\
    & \hspace{1em} character         & 1  & 248 & 13.32   & $<$~0.01 \\
    & \hspace{1em} word              & 1  & 219 & 5.74    & 0.02  \\
\midrule
\multirow{16}{*}{\shortstack{Lexical\\features}}  
    & \textit{Content words} & & & & \\
    & \hspace{1em} noun                    & 1  & 158 & 14.18   & $<$~0.01 \\
    & \hspace{1em} verb                    & 1  & 87  & 2.17    & 0.14 \\
    & \hspace{1em} adjective               & 1  & 11  & 0.74    & 0.39 \\
    & \hspace{1em} adjectival verb         & 1  & 40  & 0.95    & 0.33 \\
    & \hspace{1em} adverb                  & 1  & 22  & 0.00    & 1.00 \\
    & \textit{Lexical choice} & & & & \\
    & \hspace{1em} word frequency          & 1  & 301 & 0.54    & 0.46 \\
    & \hspace{1em} common noun             & 1  & 160 & 11.84   & $<$~0.01 \\
    & \hspace{1em} proper noun             & 1  & 6   & 0.75    & 0.39 \\ 
    & \textit{Character type} & & & & \\
    & \hspace{1em} hiragana                & 1  & 215 & 0.01    & 0.92  \\
    & \hspace{1em} katakana                & 1  & 99  & 0.00    & 1.00  \\
    & \hspace{1em} kanji                   & 1  & 212 & 5.63    & 0.02  \\
    & \hspace{1em} symbol                  & 1  & 155 & 2.20    & 0.14  \\
    & \hspace{1em} digits                  & 1  & 15  & 0.00    & 1.00  \\
\midrule                                
\multirow{7}{*}{\shortstack{Syntactic\\features}} 
    & \textit{Dependency tree} & & & & \\
    & \hspace{1em} max depth of dep. tree   & 1  & 155 & 0.28    & 0.60 \\
    & \hspace{1em} ave. depth of dep. tree  & 1  & 170 & 0.59    & 0.44 \\
    & \hspace{1em} max length of dep. links  & 1  & 201 & 0.01    & 0.91 \\
    & \textit{Others} & & & & \\
    & \hspace{1em} noun phrases            & 1  & 171 & 8.05    & $<$~0.01 \\
    & \hspace{1em} perplexity              & 1  & 316 & 14.15   & $<$~0.01 \\
\midrule
\multirow{6}{*}{\shortstack{Stylistic\\features}} 
    & \textit{Emotion} & & & & \\
    & \hspace{1em} joy            & 1  & 78  & 0.00    & 1.00  \\
    & \hspace{1em} anticipation   & 1  & 75  & 0.65    & 0.42  \\
    & \textit{Others} & & & & \\
    & \hspace{1em} textual specificity     & 1  & 11  & 0.75    & 0.39  \\
    & \hspace{1em} brackets         & 1  & 50  & 14.25   & $<$~0.01 \\
\bottomrule                                
\end{tabular}}
\caption{Results of the chi-square test for linguistic features. Df and N refer to the degree of freedom and the number of cases for each type of feature, respectively.}\label{tab:chi_square_test}
\end{table}

%% file: table/result_of_ad_text_generation.tex
\begin{table*}[th]
\centering
{\small
\begin{tabular}{l@{}cccrrr}
\toprule
\multicolumn{1}{c}{} & \multicolumn{1}{c}{\textbf{Instruction}} & \multicolumn{2}{c}{\textbf{Few-shot example}} & \multicolumn{3}{c}{\textbf{Success Rate $\uparrow$}} \\ \cmidrule(l){3-7}
\multicolumn{1}{c}{\multirow{-2}{*}{\textbf{Model}}} & \multicolumn{1}{c}{\textbf{w/ findings}} & \multicolumn{1}{c}{\textbf{positive}} & \multicolumn{1}{c}{\textbf{negative}} & \multicolumn{1}{c}{\textbf{Paraphrase}} & \multicolumn{1}{c}{\textbf{Attractive}} & \multicolumn{1}{c}{\textbf{Overall}} \\ \midrule
$\textrm{Llama2-zeroshot}$ & $-$ & $-$ & $-$ & 0.647 & 0.081 & 0.053 \\
$\textrm{GPT-3.5-zeroshot}$ & $-$ & $-$ & $-$ & 0.737 & 0.184 & 0.135 \\ 
$\textrm{GPT-4-zeroshot}$ & $-$ & $-$ & $-$ & 0.609 & 0.185 & 0.113 \\ \midrule
$\textrm{GPT-4-zeroshot-findings}$ & \checkmark & $-$ & $-$ & 0.887 & 0.314 & 0.278 \\
$\textrm{GPT-4-fewshot-pos}$ & $-$ & \checkmark & $-$ & 0.820 & 0.147 & 0.120 \\
$\textrm{GPT-4-fewshot-both}$ & $-$ & \checkmark & \checkmark & 0.842 & 0.286 & 0.241 \\
$\textrm{GPT-4-fewshot-findings-pos}$ & \checkmark & \checkmark & $-$ & 0.857 & 0.263 & 0.226 \\
$\textrm{GPT-4-fewshot-findings-both}$ & \checkmark & \checkmark & \checkmark & 0.842 & \textbf{0.366} & \textbf{0.308} \\ \midrule
HUMAN & $-$ & $-$ & $-$ & \textbf{0.940} & 0.232 & 0.218 \\ \bottomrule
\end{tabular}}
\caption{Human evaluation results for ad text generation models.}\label{tab:atg_result}
\end{table*}

%% file: table/jaccard_similarity_between_input_and_generated_ad_texts.tex
\begin{table}[t]
\centering
{\small
\begin{tabular}{lr}
\toprule
\multicolumn{1}{c}{\textbf{Model}} & \multicolumn{1}{c}{\textbf{Jaccard similarity}} \\ \midrule
GPT-4-fewshot-pos            & 0.379   \\
GPT-4-fewshot-both           & 0.358   \\ \cmidrule{1-2}
GPT-4-fewshot-findings-pos   & 0.473   \\
GPT-4-fewshot-findings-both  & 0.375   \\ \bottomrule
\end{tabular}}
\caption{\cameraready{Jaccard similarity between input and generated ad texts.}\label{tab:jaccard_similarity_between_input_and_generated_ad_text}}
\end{table}

%% file: table/analysis_of_generated_ad_texts.tex
\begin{table}[t]
\centering
{\small
\begin{tabular}{@{}l@{ }r@{\ \ \ }r@{\ \ \ }r@{}}
\toprule
& \textbf{PPL $\downarrow$} & \textbf{\#Char $\uparrow$} & \textbf{Brackets $\uparrow$} \\ \midrule
\textrm{GPT-4-zeroshot}              & 288.5 & 12.9  & 0.180 \\
\textrm{GPT-4-zeroshot-findings}     & 245.1 & 14.2  & 0.977 \\
\textrm{GPT-4-fewshot-pos}           & 248.2 & 13.0  & 0.105 \\
\textrm{GPT-4-fewshot-both}          & 284.5 & 13.6  & 0.128 \\
\textrm{GPT-4-fewshot-findings-pos}  & 206.3 & 14.0  & 0.511 \\
\textrm{GPT-4-fewshot-findings-both} & 206.6 & 14.8  & 0.699 \\ \midrule
\textrm{HUMAN}                       & 158.8 & 14.3  & 0.376 \\ \bottomrule
\end{tabular}}
\caption{Analysis results of linguistic features for generated ad texts. PPL and \#Char represent the average perplexity and the number of characters of the generated texts, respectively. Brackets represent the proportion of generated texts that contain brackets.}\label{tab:atg_analysis}
\end{table}

%% file: table/alignment_of_preference_and_pctr.tex
\begin{table}[t]
\centering
{\small
\begin{tabular}{@{ }l@{ }rrr@{ }}
\toprule
& \textbf{\#Cases} & \textbf{\#Aligned $\uparrow$} & \textbf{Ratio $\uparrow$} \\ \midrule
Entire evaluation set & 725 & 355 & 0.490 \\
\hspace{1em} high-agreement group & 316 & 173 & 0.547 \\
\hspace{1em} low-agreement group & 409 & 182 & 0.445 \\ \bottomrule
\end{tabular}}
\caption{Alignment of human preferences with pCTR. \#Aligned indicates the cases where ad texts favored by the majority of judges exhibited a higher pCTR.}\label{tab:alignment_preferece_pctr}
\end{table}

%% file: table/result_of_online_evaluation.tex
\begin{table}[t]
\centering
{\small
\begin{tabular}{lrrr}
\toprule
\multicolumn{1}{c}{\textbf{Ad group}}
& \multicolumn{1}{c}{\textbf{Imp} $\uparrow$}
& \multicolumn{1}{c}{\textbf{Click} $\uparrow$} 
& \multicolumn{1}{c}{\textbf{Cost} $\uparrow$} \\
\midrule
HR$_\mathrm{adg1}$ & 1.628 & 1.233 & 1.531 \\
Education$_\mathrm{adg2}$ & 0.912 & 0.843 & 0.740 \\
Education$_\mathrm{adg3}$ & 0.627 & 0.365 & 0.401 \\ \bottomrule
\end{tabular}}
\caption{Relative improvements of our paraphrased ads compared to original ads in online evaluation.}\label{table:online_evaluation}
\end{table}

%% file: image/instruction_for_paraprhase_creation_human.tex
\begin{figure}[t]
 \centering
  \includegraphics[width=1.0\linewidth]{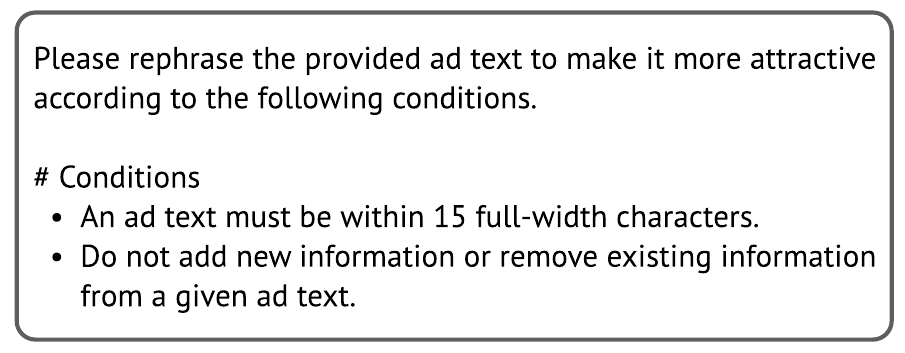}
 \caption{Instruction for paraphrase creation presented to human experts.}
 \label{fig:instruction_for_paraphrase_creation_human}
\end{figure}

%% file: image/prompt_for_paraprhase_creation_llms.tex
\begin{figure}[t]
 \centering
  \includegraphics[width=1.0\linewidth]{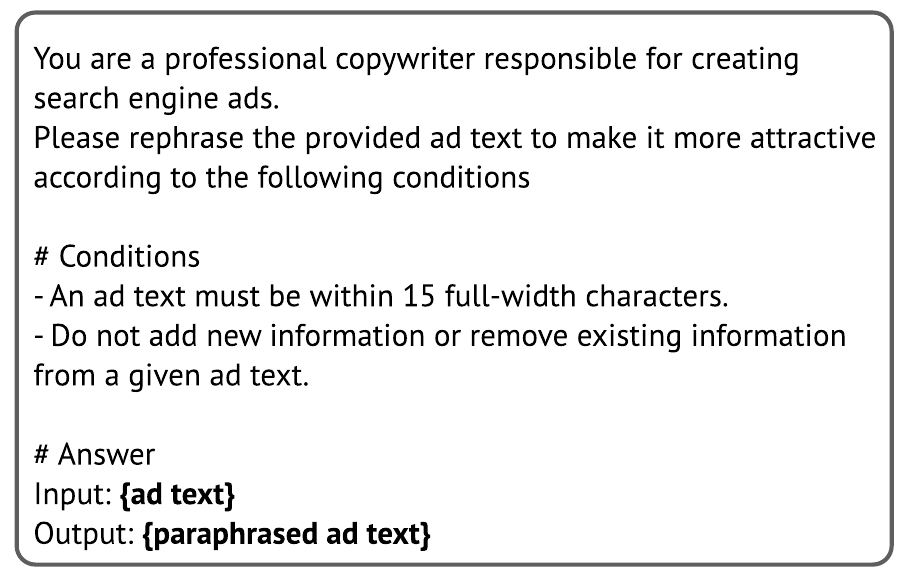}
 \caption{Prompt for paraphrase creation presented to LLMs.}
 \label{fig:prompt_for_paraphrase_creation_llm}
\end{figure}

%% file: image/paraphrase_identification.tex
\begin{figure}[t]
 \centering
  \includegraphics[width=1.0\linewidth]{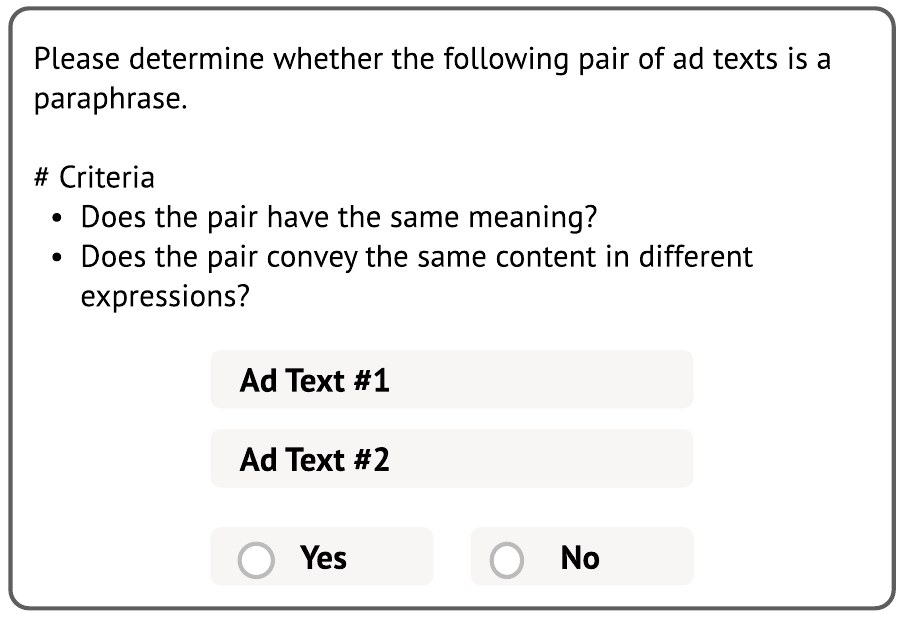}
 \caption{User interface for paraphrase identification.}
 \label{fig:user_interface_paraphrase_identification}
\end{figure}

%% file: image/paraphrase_and_non_paraphrase_examples_for_paraphrase_identification.tex
\begin{figure}[t]
 \centering
  \includegraphics[width=1.0\linewidth]{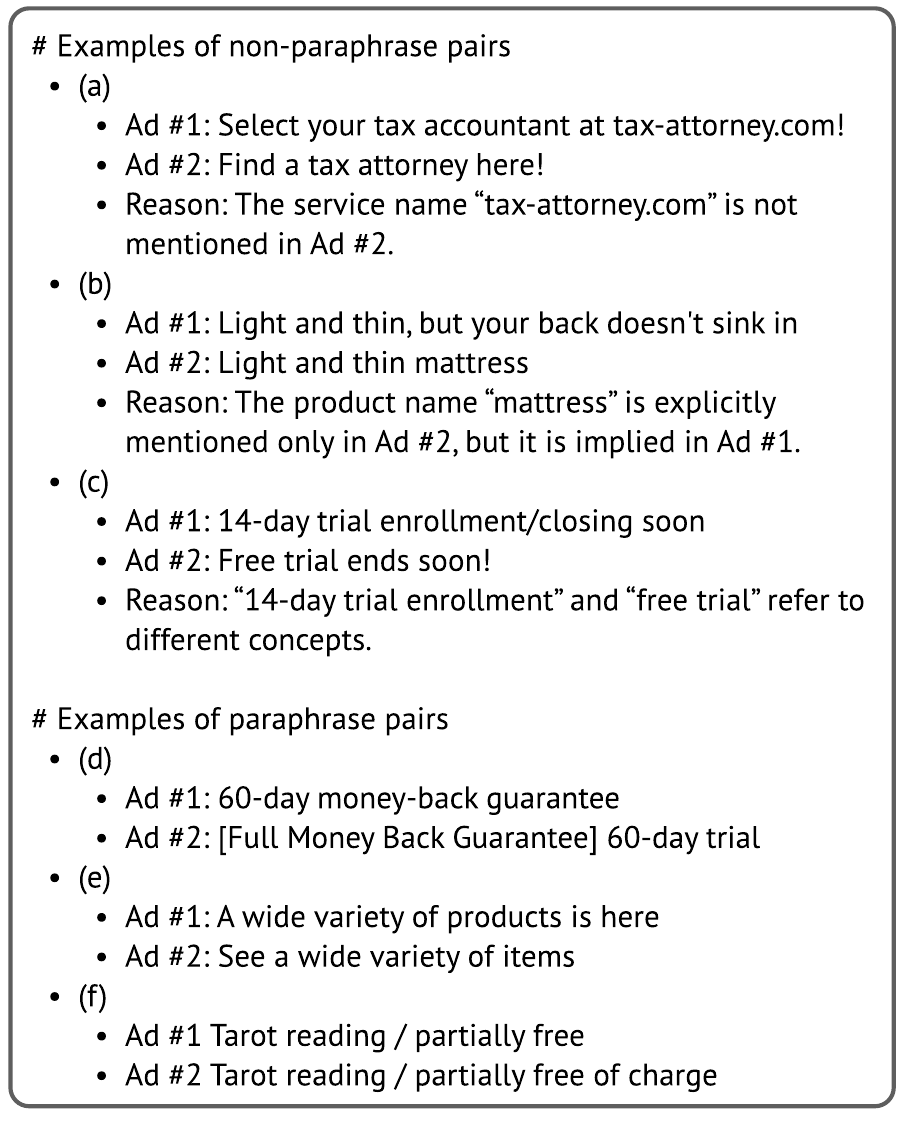}
 \caption{\cameraready{Annotation examples presented to judges for paraphrase identification.}}
 \label{fig:annotation_example_for_paraphrase_identification}
\end{figure}

%% file: image/attractiveness_evaluation.tex
\begin{figure}[t]
 \centering
  \includegraphics[width=1.0\linewidth]{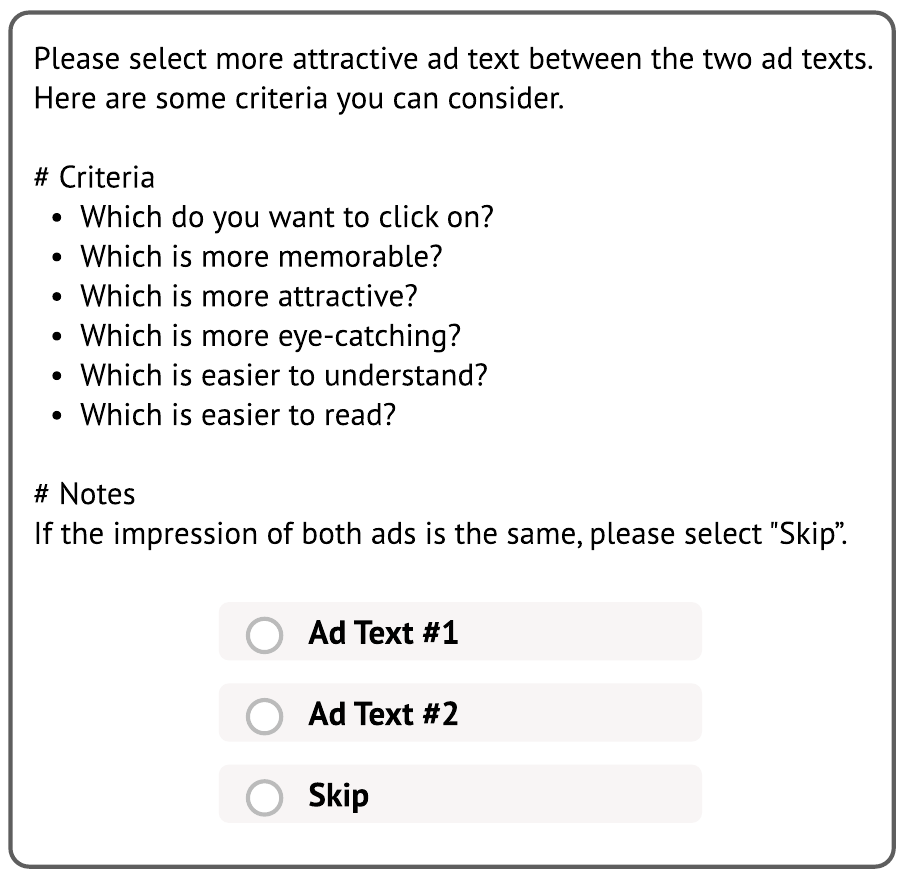}
 \caption{User interface for attractiveness evaluation.}
 \label{fig:user_interface_attractiveness_eval}
\end{figure}

%% file: image/prompt_for_ad_text_generation.tex
\begin{figure}[t]
 \centering
  \includegraphics[width=1.0\linewidth]{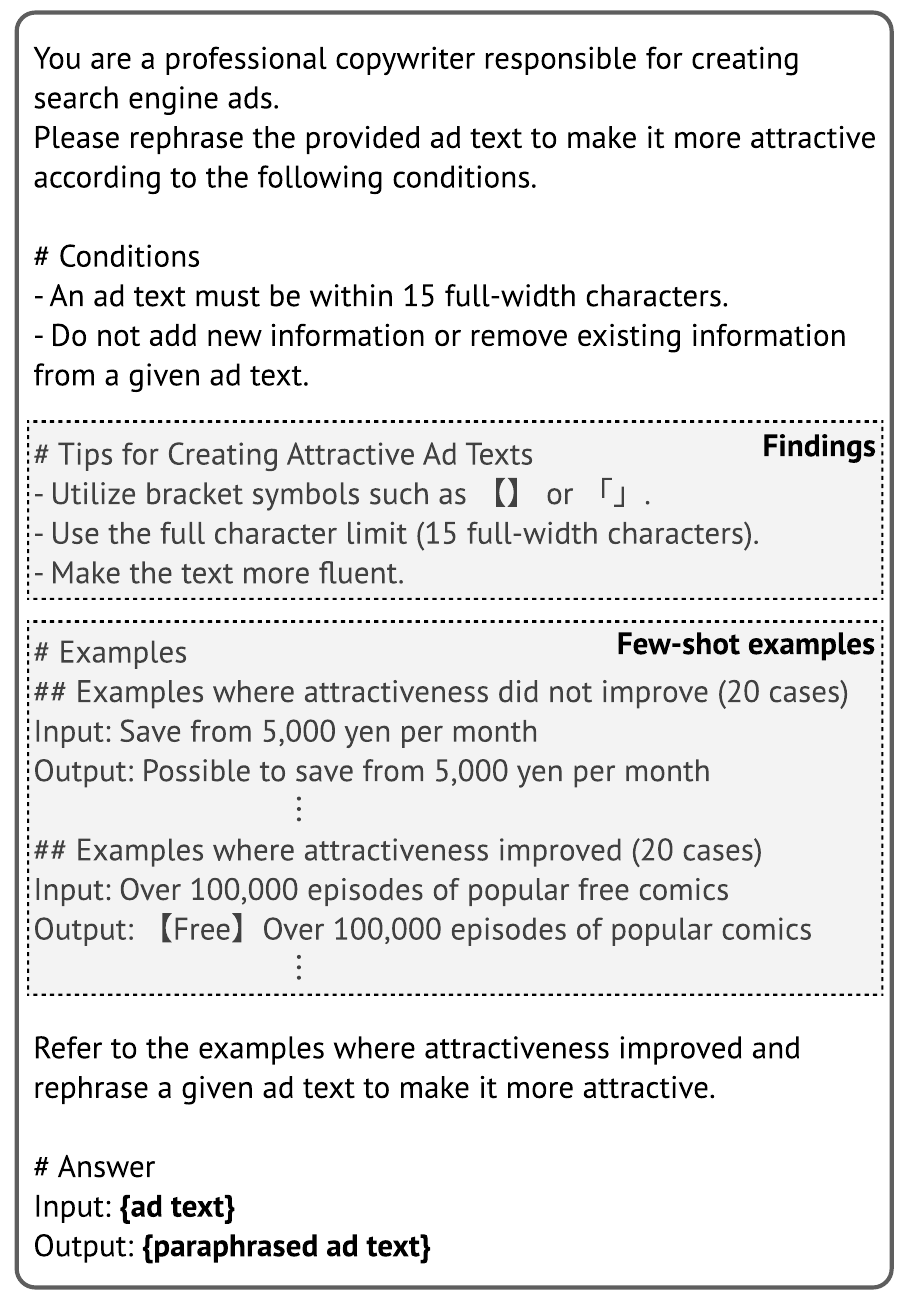}
 \caption{Prompt for ad text generation experiment. For visibility, Japanese prompt is translated into English.}
 \label{fig:prompt_for_ad_text_generation}
\end{figure}